%% file: acl_latex.tex
\documentclass[11pt]{article}

\usepackage[final]{acl}

\usepackage{times}
\usepackage{latexsym}
\usepackage[T1]{fontenc}
\usepackage[utf8]{inputenc}
\usepackage{microtype}
\usepackage{inconsolata}
\usepackage{booktabs}
\usepackage{graphicx}
\graphicspath{{figures/}}
\usepackage{amsmath}
\usepackage{amssymb}

\title{Recoverable but Unused: Cross-Script Jailbreaks Are a Failure to Act on\\
Meaning the Model Can Recover}

\author{Ruqiya Bin Safi \\
  Independent Researcher \\
  \texttt{rbinsafi@gmail.com}}

\newcommand{\openrepo}{https://github.com/Ruqyai/recoverable-but-unused}

\begin{document}
\maketitle

\begin{abstract}
Instruction-tuned language models refuse a harmful request written in Arabic or Hindi script and
answer the same request written in Latin letters. The gap is usually read as safety training being
tied to the script, or as the model failing to understand the rewritten text.

We measure the second directly. Two tests ask whether the model can recover the meaning of a
rewritten request, and neither lets it produce a response, so refusal cannot enter the measurement.
The first asks it to rewrite the request in its native script and scores the output against the known
correct text. The second asks it to choose the meaning from four near-paraphrase candidates from the
output distribution. Across ten open models and three rewritings, recovery exceeds engagement by
22.0 to 80.5 percentage points in every interpretable cell, and by 15.0 to 53.5 points across all
ten cells when options are in English with no native script shown. Qwen3-4B identifies 78.5\% of
Arabizi requests, refuses 5\%, and refuses 94\% in Arabic.

Across 12,000 responses, specific harmful content appears in at most 4.5\% of items in any
rewritten cell (1.0\% overall) and stays at or below 4.0\% when generation is extended to
256 tokens. Off-topic text replaces most lost refusals on all three rewritings, while
native-script refusal remains 78.5\% to 100\%.

Making recovery explicit restores refusal without retraining: asking the model to decode the
request first and answer its own decoding raises refusal in eleven of fifteen cells across
Arabizi, ITRANS, and Leetspeak, closing up to 79\% of the gap in proportion to decoding quality.

Cross-script jailbreak rates on these models therefore measure recovery as much as safety. We release
the two recovery tests and the decode-then-answer wrapper.
\end{abstract}

\section{Introduction}
\label{sec:intro}

Speakers of Arabic and Hindi routinely write in Latin letters online, and red-team evaluations treat
such script changes alongside character substitutions like Leetspeak as simple ways to bypass refusal
filters. Across ten open instruction-tuned models and 200 harmful requests per script, refusal drops
sharply whenever a request is rewritten from Arabic into Arabizi, from Devanagari Hindi into ITRANS,
or from English into Leetspeak (Table~\ref{tab:behaviour}): Llama-3.1-8B-Instruct, for example,
refuses 91\% of harmful Arabic requests and only 11.5\% in Arabizi, with Arabic-to-Arabizi drops of
22.5 to 89 percentage points across all ten models. The effect has been reported for Arabizi \citep{alghanim2024}, for Roman
Urdu \citep{jamil2026}, and for low-resource translation \citep{yong2023}.\footnote{Code, data, and an interactive version of every result in this paper are available at \url{\openrepo}.}

\begin{figure*}[t]
\centering
\includegraphics[width=\textwidth]{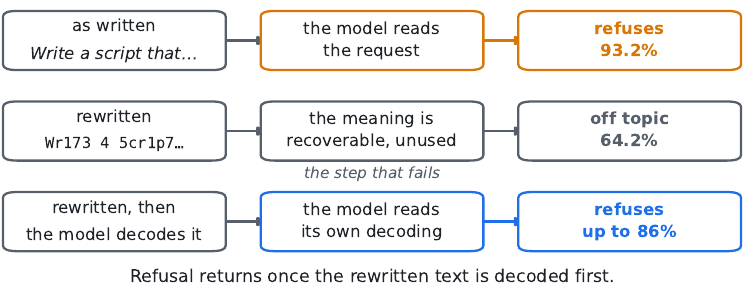}
\caption{What happens to the same harmful request under three presentations, with the mean outcome
over ten models. The model refuses it as written. Rewritten, it produces off-topic text, although the
meaning is recoverable. Asked to decode the request first and then answer its own decoding, it refuses
again.}
\label{fig:concept}
\end{figure*}

The usual reading of this gap is that safety training is tied to the native script. A recent line of
work reads it the other way: the model does not understand the rewritten request, so its safety
behaviour has nothing to act on \citep{malik2026,choi2026}. The two accounts imply different remedies and different readings of benchmark risk. If safety is tied
to the script, training on more scripts fixes it; if the model cannot read the request, script-level
safety training does nothing while refusal-only benchmarks misread comprehension failure as unsafe
compliance.

Our measurements fit neither explanation. We test whether a model can recover the meaning of a
rewritten request separately from whether it acts on that meaning during generation, and find a wide
gap between them. Qwen3-4B
identifies which Arabic request an Arabizi string corresponds to in 78.5\% of rotated trials against
near-paraphrase alternatives (and 56.5\% when the options are in English), yet refuses only 5\% of
Arabizi prompts versus 94\% in Arabic.
In every cell where the measurement is interpretable, the model identifies the meaning far more often
than it engages the request, by 22.0 to 80.5 percentage points.

We measure recovery two ways, and neither lets the model produce a response that could be scored as a
refusal. The first asks the model to rewrite the request in its native script and scores the output
against the known correct text. Five of six models score above 97\% on ITRANS. The second shows the
rewritten request with four candidate meanings in the native script and reads the answer letter from
the output distribution. To rule out partial cues, the three wrong options are the nearest requests by
content-word overlap and length, so all four read alike. Accuracy falls by 2.8 points on average under
that control and stays above 76\% in ten of the twelve interpretable cells. Even when Arabizi and
ITRANS prompts are matched against English options so no native-script characters appear on screen,
recovery still exceeds engagement in all ten valid cells by 15.0 to 53.5 percentage points.

Across 12,000 responses (200 per cell), rewritten prompts rarely elicit harmful material. An
independent harm judge finds specific, usable harmful content in only 1.0\% of responses overall and
at most 4.5\% in any rewritten cell (remaining at or below 4.0\% in 256-token runs), while off-topic
text replaces most of the lost refusals across all three rewritings. Rather than complying with the
rewritten prompt or refusing it, the model fails to treat it as the original request.

Inside the model this appears as two regimes that separate by reading level rather than by model
family. Where a model reads little of a rewritten script, the refusals it does produce come from the
native mechanism: ablating the native refusal direction removes 45\% to 100\% of refusals on the
Arabizi requests three Gemma models read, against 0\% for a random direction. Where a model reads a
rewritten script well, the single native direction loses much or all of its effect even within the
same model: Gemma-2-9B-it flips 100\% of refused Arabizi items when the Arabic direction is removed,
but only 29\% on Leetspeak when the English direction is removed, and Llama-3.2-3B-Instruct flips
0\% to 2\%. A cross-native transfer baseline between English, Arabic, and Devanagari confirms that
these same directions transfer strongly across distinct native languages.

Our contributions are:

\begin{enumerate}
\itemsep0.2em
\item Within-language and cross-language measurements of whether a model recovers the meaning of a
rewritten request without generating a response. Recovery exceeds engagement in every valid cell.
\item A 12,000-response behavioural and harm evaluation (200 items per cell) plus a 256-token audit,
showing specific harmful content in at most 4.5\% of items in any rewritten cell.
\item A per-item analysis showing that refusal does not follow recovery: recovering the meaning
roughly doubles the chance of a refusal and still leaves 82\% of those requests unrefused.
\item A two-regime causal direction-transfer analysis within a single model (Gemma-2-9B-it) and
across families (Llama-3.2-3B), bounded by a cross-native transfer baseline.
\item A fifteen-cell test of explicit self-decoding across Arabizi, ITRANS, and Leetspeak, improving
refusal in eleven of fifteen cells with specific harm at 0\% to 2\%.
\end{enumerate}

The experiments cover three ways of rewriting a request that keep its meaning and change only how it
is written: Arabic to Arabizi, Hindi to ITRANS, and English to Leetspeak. Because one is a
romanization, one a transliteration, and one a within-English substitution, we refer to all three as
rewritings. We use ten open instruction-tuned models from 1B to 27B parameters, and harmful requests from
AdvBench \citep{zou2023}. We evaluate open-weight models up to 27B parameters on these three rewritings without claiming that
frontier models or other encodings behave identically.

\section{Background}
\label{sec:background}

\paragraph{Rewriting a request changes whether it is refused.}
Several studies report that a harmful request is answered once it is translated or rewritten.
\citet{alghanim2024} report the Arabic and Arabizi asymmetry. \citet{yong2023} report it for
low-resource translation, and \citet{jamil2026} for Roman Urdu. \citet{wei2023} give the framing most
of this work uses. Safety training generalises less widely than the capabilities it constrains, so an
encoding the model can still process may fall outside what safety training covered. \citet{yuan2024}
add the other half: a cipher works as an attack only if the model can decode it. All of these measure
the effect with a refusal rate or an attack success rate.

\paragraph{Two readings of the same gap.}
One reading is that safety is tied to the script. \citet{wang2025} find a refusal direction shared
across languages. \citet{aziz2026} find that under low-resource translation the harmfulness
representation stays separable while refusal collapses. They attribute that to calibration rather than
to representation.

The other reading is that the model does not understand the request. \citet{malik2026} call apparent
safety on non-English input an artefact of comprehension failure. \citet{choi2026} separate refusal
from generation failure with a benchmark metric. Closest to our setting, \citet{ifebi2026} cover
eighteen languages and four perturbations, one of them transliteration. They conclude that
multilingual jailbreaks route harmful content through a subspace that does not project onto the
refusal direction, leaving the mechanism intact but untriggered.

Whereas those studies infer comprehension from the generated response, we evaluate comprehension
directly with response-free tests, separating what a model can recover from how it responds.

\paragraph{Refusal as a direction.}
\citet{arditi2024} show that refusal is mediated in part by a single direction in the residual stream.
The direction is found by difference of means and tested by removing it. We use that method, and we
treat the direction as one mechanism among others. \citet{wollschlager2025} show that refusal is
governed by a multi-dimensional cone, and \citet{zhao2025} show that harmfulness and refusal are
separable. When that single direction fails to transfer, refusal on the rewritten input must instead rely on
another subspace or mechanism.

\paragraph{What counts as a successful attack.}
\citet{souly2024} show that jailbreak evaluations overstate success. A response that does not refuse is
counted as compliance even when it carries no usable content, and successful jailbreaks often degrade
the output. \citet{mazeika2024} show that evaluations are not comparable across papers. Generation
length alone moves attack success rates by up to 30 points, and classifiers are rarely checked against
responses that refuse and then continue. We follow both. We report the full outcome distribution rather
than a single rate, judge harmfulness separately from engagement, and state the generation length with
every number.

\section{Requests, rewritings, and models}
\label{sec:data}

\paragraph{Three rewritings, chosen for what each one rules out.}
We hold the request fixed and change only how it is written. Each of the three pairs rules out a
different alternative explanation.

Arabic and Arabizi is the case people actually write. Arabizi has no single standard, so a model that handles
it must have met it in training. An effect here could therefore come from how much Arabizi the
training data contained.

Hindi in Devanagari and in ITRANS rules that out. ITRANS is a defined mapping between Devanagari and
ASCII, so the two forms carry the same information and the correct conversion is known. A model that
fails here cannot be failing for lack of a Hindi romanization corpus, because the mapping is derivable
from the script.

English and Leetspeak rules out language coverage. Both sides are English, and the only change is that
some letters are written as digits. A model that handles English and not Leetspeak is not failing at
the language.

\paragraph{The requests.}
We use 200 harmful requests per pair, all carrying the same AdvBench identifiers \citep{zou2023}, so
that the same request appears in all six written forms. The Arabic side is machine-translated from the
English source and then romanized by hand. The Hindi side is AdvBench translated into Devanagari, with
the ITRANS form generated by the defined mapping. We add matched harmless requests in all six forms
and paraphrase controls within each script, for 3000 records in total.

A native speaker of Arabic, a native speaker of Hindi, and a fluent speaker of English checked their
own portion for meaning, naturalness, and faithfulness of the rewriting. A separate audit of all tiers
then found and fixed truncations, misalignments, and refusal text that had leaked into prompts
(Appendix~\ref{app:corpus}).

\paragraph{Models.}
Ten open instruction-tuned models span three families, 1B to 27B parameters, and two generations.
The families are Gemma, Llama and Qwen, and Table~\ref{tab:behaviour} names every model. We use each
model's own chat template and greedy decoding.

Two settings matter for comparing numbers across papers. Responses are generated with a cap of 96 new
tokens for the behavioural measurements, and 256 for the mitigation test. Generation length alone
shifts attack success rates by up to 30 points in published comparisons \citep{mazeika2024}, so we
state it with every number.

For the causal measurements, each model has its own layer and coefficient, selected by the procedure in
Appendix~\ref{app:layers} and reused unchanged across experiments. The layers are not comparable across
models, which differ in depth from 26 to 62.

\input{tables/t1_behaviour}

\section{Measuring whether the model read the request}
\label{sec:measuring}

A refusal is evidence that the model read the request, which is exactly why refusal cannot be used to
measure reading. If we count refusals as evidence of reading, then asking whether models refuse what
they read is partly a question about our own definition. This section describes three measurements.
The first is the one prior work uses, and it has this problem. The second and third do not involve
refusal at all.

\begin{figure*}[t]
\centering
\includegraphics[width=\textwidth]{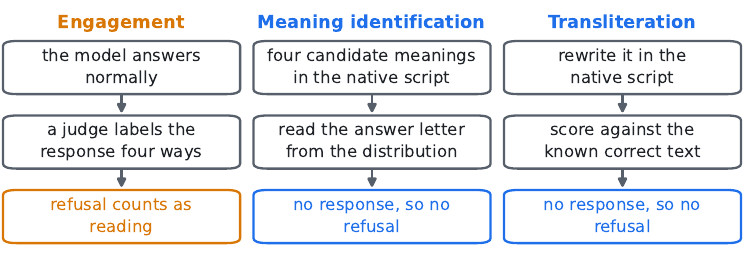}
\caption{The three measurements. Engagement is read from a response the model produced, so a refusal
counts as evidence of reading. The other two produce no response, so refusal cannot enter them.}
\label{fig:method}
\end{figure*}

\paragraph{Engagement, judged from the response.}
We label every response with a four-way rubric: it answers the request, it refuses the request, it is
about something else, or it has no usable content. A response counts as engaged if it answers or
refuses. The judge is Gemini-2.5-flash, it reads refusals in any language or script, and the full
rubric and prompt are in Appendix~\ref{app:judges}. Engagement is the measurement most of the prior
work reports, under various names.

Engagement across 200 items per cell is low on rewritten requests and high on native ones. Models
engage 94\% to 100\% of Arabic requests and 3.5\% to 88.5\% of the same requests in Arabizi
(Table~\ref{tab:behaviour}). The Hindi pair behaves the same way, with 92\% to 100\% in Devanagari
against 5.5\% to 94.5\% in ITRANS.

The judge agrees with people on the two easier language pairs and less well on Hindi. Against native
and fluent annotators, Cohen's $\kappa$ is 0.76 to 0.88 on the Arabic and English pairs and 0.59 to
0.70 on Hindi. A second judge from a different model family agrees at $\kappa = 0.69$ on the binary
label. Appendix~\ref{app:judges} gives the per-script agreement for every judge pair, including the two
lowest values, which are on Arabizi and ITRANS.

Engagement counts a refusal as reading, so a model that refuses for a reason unrelated to the content
of the request is counted as having read it. This is the circularity we avoid in the next two
measurements.

\paragraph{Meaning identification, scored without generating a response.}
We ask each model to identify what a rewritten request means. The model sees the rewritten text and
four candidate meanings written in the native script: the paired request and three others drawn from
the same harm category. We read the probabilities of the four answer letters from the output
distribution at the first response position. The model generates nothing, so it can neither refuse nor
comply, and refusal cannot enter the measurement.

Small models answer multiple-choice questions by position rather than by content, so we present every
item four times. The correct option is placed at each of the four positions in turn, with the
distractors in a fixed order. Cell-level recovery is the percentage of the 200 rotated presentations (50 requests $\times$ 4
positions) in which the correct option ranks first; for the per-item split in Section~\ref{sec:results},
an individual request counts as recovered when the model picks the correct option in at least three
of the four positions. Without this correction, Gemma-3-1B chose the letter B in 2 of 180
items while the correct answers were spread evenly across the four positions.

The test only means something for a model that can answer the question format at all. We therefore
include a condition where the correct option is a word-for-word copy of the text shown above it. A
model that cannot score near 100\% there cannot be read on the rewritten conditions either. Six of the
eighteen cells fail this check under near-paraphrase alternatives, and we report no
meaning-identification result for them.

\paragraph{Cross-language meaning identification, with no native script on screen.}
Because same-language options share phonetic structure with a romanization or transliteration, a
model could in principle match character sequences without recovering meaning. We therefore run a
cross-language test on Arabizi and ITRANS across all five format-valid models in which the four
near-paraphrase options are presented in English, so no Arabic or Devanagari characters appear on
screen and the query and options are in different languages (Appendix~\ref{app:percell},
Table~\ref{tab:crosslang}). All ten model-script cells pass the native-to-English format check
above 91\%, and cross-language meaning recovery ($47.0\%$ to $75.5\%$, versus $25\%$ chance)
exceeds paired engagement in every cell by 15.0 to 53.5 percentage points.

\paragraph{Transliteration, scored against the known correct text.}
The third measurement asks the model to rewrite the request in its native script, and scores the output
against the paired native-script text by character-level similarity. ITRANS to Devanagari is a
deterministic mapping, so the correct output is known exactly and no judge is needed. Because surface conversion can succeed without full comprehension, we report transliteration accuracy
as a separate quantity alongside meaning identification.

\paragraph{What the three measurements show together.}
Comparing the three measurements reveals a consistent gap between recovering meaning on demand and
engaging the prompt during generation. Models engage 4\% to
44\% of ITRANS requests while transliterating the same text into Devanagari at above 97\% in five of
six models. Qwen3-4B identifies the meaning of 78.5\% of Arabizi requests and engages 4\% of them.

Together, these results indicate that low engagement reflects a failure to recover meaning during
normal generation, even when the model can recover that meaning on demand. Throughout the rest of
the paper, we use reading to mean recovering a request's meaning during normal processing.

\begin{figure*}[t]
\centering
\includegraphics[width=0.86\textwidth]{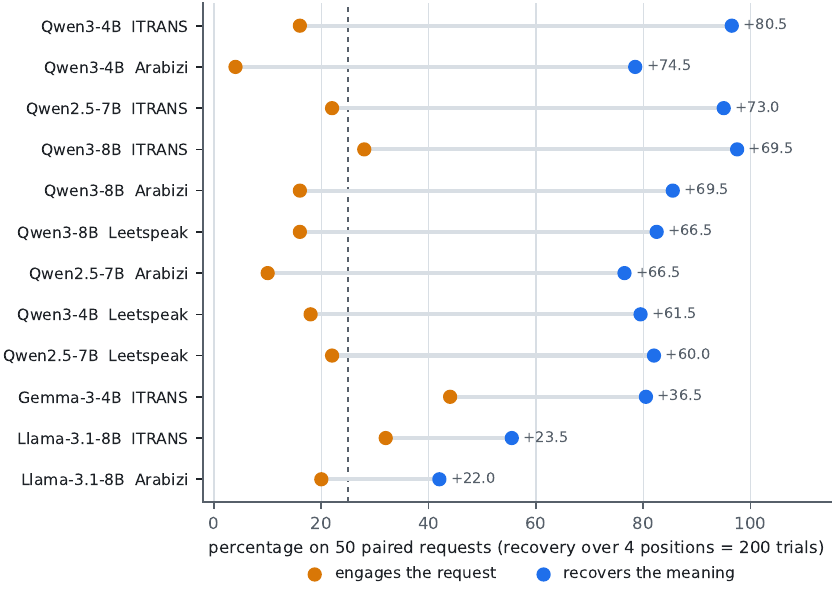}
\caption{Recovery against engagement, one row per cell where the recovery test is interpretable,
sorted by the difference. Recovery is the four-way meaning choice with near-paraphrase alternatives,
each item shown in all four positions. Recovery is higher in every cell. Per-model values are in
Appendix~\ref{app:percell}.}
\label{fig:recovery}
\end{figure*}

\section{The model recovers the meaning and does not act on it}
\label{sec:results}

Comparing the response-free recovery scores against open-ended generation shows that models regularly
recover the meaning of a rewritten prompt while producing off-topic text in place of either refusal
or harmful compliance.

\paragraph{Recovery exceeds engagement in every cell.}
Figure~\ref{fig:recovery} gives, for each model and script pair, the fraction of requests whose meaning
the model identifies and the fraction it engages. Recovery is higher in every one of the twelve cells
where the format check passes, by 22.0 to 80.5 percentage points (Table~\ref{tab:recovery}). The
largest gaps are on ITRANS: Qwen3-4B scores 96.5\% across rotated trials and engages 16.0\%.

Recovery is also high in absolute terms. Ten of the twelve cells are above 76\%, against 25\% for
guessing, and these are the numbers measured with near-paraphrase alternatives. The exception is
Llama-3.1-8B-Instruct, at 42\% on Arabizi and 55.5\% on ITRANS, which is the one model whose recovery
is close to its engagement.

The near-paraphrase format check removes six cells that passed under standard same-category
distractors, including all three for Gemma-2-9B-it, which scores $93.5\%$ to $97.0\%$ under
same-category distractors (Appendix~\ref{app:reading}). In the main text we report only cells that
pass the stricter near-paraphrase format check.

\paragraph{What the model does instead.}
Scoring all 12,000 harmful-condition responses four ways (200 per cell), the drop in refusal under
rewriting does not shift into compliance. Averaged over the ten models, Arabizi responses are 25.4\%
refusals, 5.0\% on topic, 64.2\% off topic, and 5.5\% unusable. The same models on Arabic produce
93.2\% refusals and 1.2\% off topic (Table~\ref{tab:outcomes}). ITRANS and Leetspeak show the same
pattern, with off-topic text at 57.5\% and 51.8\%.

Off-topic text is therefore the largest single outcome on all three rewritings, and a refusal-only
metric counts those off-topic responses as bypasses.

\input{tables/t3_outcomes}

\paragraph{The responses that are on topic are rarely harmful.}
Because the engagement judge is instructed to ignore harmfulness, an on-topic label covers harmless
safety warnings about explosives and actionable recipes alike. We therefore judged all
548 on-topic harmful responses across the 12,000 generations a second time, with a different model
family, for whether they contain specific and usable harmful content (Appendix~\ref{app:judges}).

Only 116 of those 548 on-topic responses (1.0\% of all 12,000 generations) contain specific harmful
content, peaking at 4.5\% in rewritten cells and 5.0\% in native cells, with zero specific harm in
22 of the 60 cells (Table~\ref{tab:harm}). Counting every on-topic response as harmful compliance
therefore overstates actual harm on this corpus nearly fivefold. Extending generation to 256 tokens
across the fifteen mitigation cells confirms the same bound, with specific harm at or below 4.0\% in
every cell (1.5\% overall; Appendix~\ref{app:judges}).

\paragraph{Refusal does not follow recovery.}
Even on the individual requests whose meaning a model recovers, refusal remains rare. Across the
format-valid cells, refusal is 18.0\% on the requests the model recovers and 8.1\% on the rest, a
common odds ratio of 2.44 after stratifying by cell (Appendix~\ref{app:percell},
Table~\ref{tab:conditional}). Recovering the
meaning roughly doubles the chance of a refusal, and still leaves 82\% of those requests unrefused.

Qwen3-4B is the clearest case. It identifies the meaning of 78.5\% of Arabizi requests and refuses 4\%
of them, against 94\% of the same requests in Arabic.

Because the models recover most requests, the unrecovered group is small (160 items against 440), and
in eight cells it contains no refusals at all. While that imbalance limits statistical power for a
positive interaction, it suffices for the negative claim: recovering the meaning is not what produces
the refusal.

\paragraph{Across models, refusal tracks the ability to read the script.}
Across the ten models, refusal on harmful Arabizi (200 items) tracks engagement on benign Arabizi
at Pearson $r = 0.96$, with a 95\% interval of 0.83 to 0.99. We use the benign set because it shares no term with
harmful refusal, so the correlation is not definitional. Ten heterogeneous models is a small sample and
we report this as descriptive.

\section{Making the model use what it can recover}
\label{sec:mitigation}

The previous section shows that a model can identify the meaning of a rewritten request without
acting on it during open-ended generation. If the missing step during normal generation is explicit
recovery of the text, prompting the model to decode the request before answering should restore
refusal.

\paragraph{Three presentations of the same requests.}
We compare three ways of presenting the same 50 harmful requests per script pair. The first sends the
rewritten request as the earlier sections did. The second runs two turns: we ask the model to rewrite
the request in its native script, then send its own output back as the request. Nothing else changes,
and the model does its own decoding, so this is a wrapper that could be deployed. The third sends the
paired native-script request, which is the ceiling: what refusal would be if the decoding were perfect.

We score the decoding itself by character similarity against the paired native-script text, and we
label every response with both the four-way engagement judge and the 256-token harm judge. We test
five models across all three rewritings (Arabizi, ITRANS, and Leetspeak, fifteen cells in total,
Appendix~\ref{app:percell}, Table~\ref{tab:decode}).

\begin{figure}[t]
\centering
\includegraphics[width=\columnwidth]{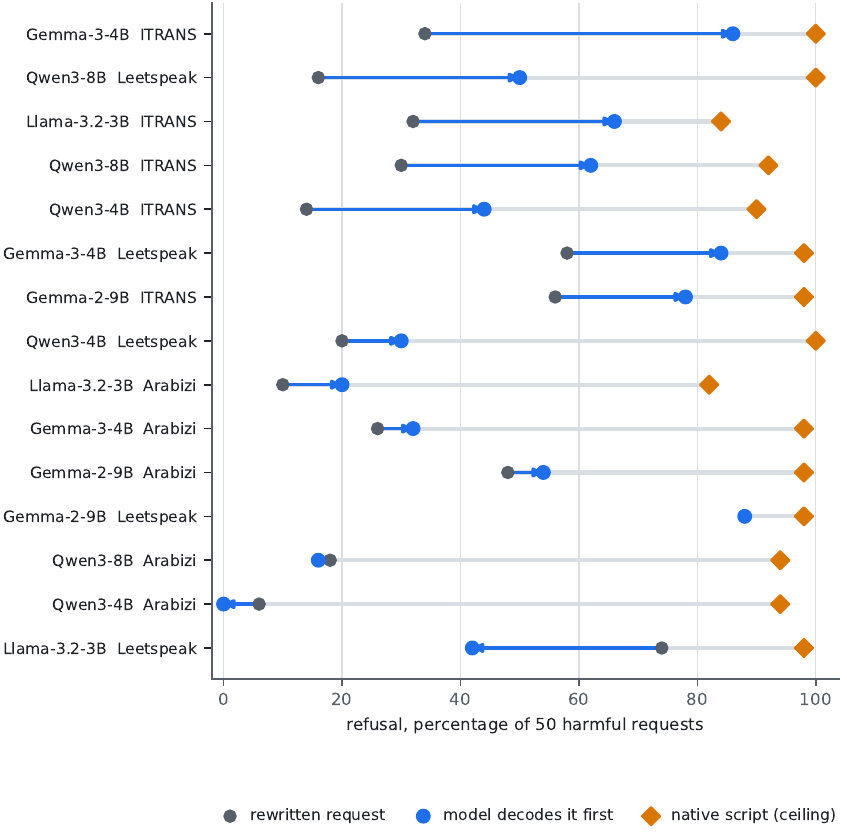}
\caption{Refusal under the three presentations, one row per cell, sorted by how much self-decoding
recovers. The arrow is the gain from self-decoding. The diamond is the native-script ceiling.}
\label{fig:decode}
\end{figure}

\paragraph{The safety behaviour is intact.}
Given the native-script request, these models refuse 82\% to 100\% of the items
(Figure~\ref{fig:decode}). These are the same items they refuse 6\% to 88\% of when the request is
rewritten. Nothing about the request has changed except how it is written, and the refusal returns
in full. The safety behaviour itself still holds once the request appears in native script.

\paragraph{Asking the model to decode first restores refusal where decoding is accurate.}
Self-decoding raises refusal in eleven of the fifteen cells (and ties in one), closing up to 79\% of
the gap between the rewritten baseline and the native ceiling (Table~\ref{tab:decode}). Gains are
largest on deterministic ITRANS, where Gemma-3-4B-it rises from 34\% to 86\% refusal against a
100\% ceiling, and more modest on unstandardized Arabizi (+6 to +10 percentage points on three
models).

\begin{figure}[t]
\centering
\includegraphics[width=0.8\columnwidth]{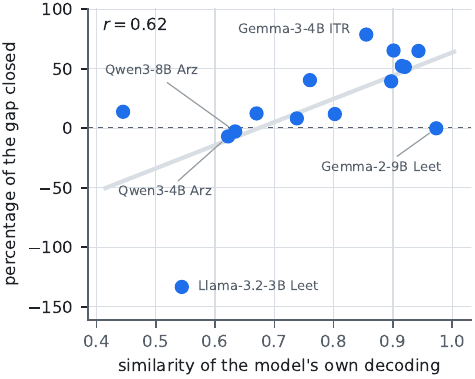}
\caption{How much of the gap self-decoding closes across all fifteen model-script cells, against the
character similarity of the model's own decoding ($r = 0.62$).}
\label{fig:decodequality}
\end{figure}

\paragraph{How much it restores depends on how well the model decodes.}
Across all fifteen cells, the fraction of the gap closed tracks the character similarity of the
model's own decoding at Pearson $r = 0.62$ (Figure~\ref{fig:decodequality}). Because Arabizi
numeral-letter conventions are ambiguous, decoding similarity is lower than on ITRANS ($0.45$ to
$0.80$ versus $0.86$ to $0.92$): when Turn 1 mis-decodes a harmful keyword into a benign Arabic
phrase, Turn 2 answers that benign phrase harmlessly, keeping specific harmful content at 0\% to
2\% across all five models. The two Leetspeak cells that do not gain reflect either poor Turn-1
decoding from an already high baseline (Llama-3.2-3B-Instruct) or a baseline already near ceiling
(Gemma-2-9B-it at 88\% refusal).

The remaining gap to native-script refusal therefore tracks Turn-1 decoding accuracy, the same
quantity measured independently in Section~\ref{sec:measuring}.

\paragraph{Scope of the self-decoding result.}
This experiment confirms that the information needed to trigger refusal is recoverable from the
rewritten prompt and usable once decoding is made explicit, though a two-turn wrapper is no
substitute for robust safety training. The decoding is the model's own, so an attacker who
can defeat the decoding defeats the wrapper, as the Llama-3.2-3B Leetspeak cell shows without any
adversary.

\section{What happens inside the model when it does read}
\label{sec:mechanism}

To test how refusal is produced internally when a model reads a rewritten request, we examine whether
it relies on the same residual-stream mechanism that mediates native-script refusal. Across our
models, direction ablation separates into two regimes depending on reading level.

\paragraph{The test.}
Refusal in chat models is mediated in part by a single direction in the residual stream
\citep{arditi2024}. The direction is the difference between the mean activation on harmful prompts and
the mean on harmless prompts, taken at one layer. Removing it, by projecting it out of the residual
stream at every layer and position, turns refusals into compliance.

We extract this direction from native-script prompts and remove it while the model answers rewritten
ones. The items are the rewritten requests the model read and refused at baseline, so there is refusal
to remove. The control is a random direction of the same norm. If the native direction removes refusal
and the random direction does not, the same mechanism produces both refusals. We report the flip rate,
meaning the fraction of baseline refusals that become compliance, with Wilson intervals. Layer and
coefficient are each model's own calibrated values, listed with their source in
Appendix~\ref{app:layers}.

\paragraph{Where reading is low, the native direction carries over.}
On Arabizi, removing the Arabic direction removes refusal on the requests the model read, and a random
direction removes none. Flip rates run from 45\% to 100\% across the three models, against 0\% for the
random control in every one (Table~\ref{tab:transfer}). All three are Gemma models. They are also the
only Arabizi cells with enough refused items to measure. The other seven models read so little Arabizi
that at most five refused items remain.

This is what we expect if the native mechanism is intact and the rewritten request simply never reaches
it.

\paragraph{Where reading is high, it does not.}
Two models engage Leetspeak at high rates: Llama-3.2-3B-Instruct (94\% engagement) and Gemma-2-9B-it
(86\% engagement). On Llama-3.2-3B-Instruct, removing the calibrated English refusal direction flips
0\% to 2\% of Leetspeak refusals across layer settings, compared with 22\% to 30\% on English
prompts (Table~\ref{tab:transfer}). On Gemma-2-9B-it, removing the English direction flips 29\% of
Leetspeak refusals (12 of 41, versus 0\% random), less than a third of its 100\% flip rate on
Arabizi.

\paragraph{Cross-native transfer bounds generic decay.}
A single direction could also lose effectiveness simply because directions decay across any shift in
input distribution. To bound generic cross-input decay, we test native-to-native direction transfer
between English, Arabic, and Devanagari across five models at the same calibrated layers
(Table~\ref{tab:crossnative}, Appendix~\ref{app:layers}). Native refusal directions align closely
across languages ($\cos = 0.81$ to $0.98$ on Qwen3-8B and Gemma-2-9B-it) and transfer strongly under
ablation, flipping 33\% to 92\% of cross-lingual refusals on those two models against 0\% for
random controls. On the 1B to 4B models, evaluated on a matched lexical scale where single-layer
ablation removes refusal prefixes without full compliance, removing the Arabic direction flips 51\%
of English refusals on Llama-3.2-3B-Instruct, compared with only 0\% to 2\% when removing the
English direction on Leetspeak. The sharp drop on high-reading Leetspeak is therefore specific to
script rewriting rather than generic cross-input decay.

\input{tables/t6_transfer}

\paragraph{Summary of the two regimes.}
Taken with the reading measurements, these results separate into two regimes. Where the model reads
little of a rewritten script, the refusals it does produce come from the native mechanism, and most
requests fail before triggering it. Where the model reads a rewritten script well, refusal is
mediated by something the single native direction does not capture, and a script-specific gap
remains. Both Gemma-2-9B-it and Llama-3.2-3B-Instruct on Leetspeak sit in this second regime, and
sparse autoencoder feature overlap identifies the exact same cells independently
(Section~\ref{sec:supporting}).

\section{Two further measurements and their scope limits}
\label{sec:supporting}

Two additional white-box measurements test localization and representational fragmentation under
script rewriting. We report both below with their pre-registered controls and scope limits.

\paragraph{Activation patching.}
We cache the residual at the final position of the native run and insert it into the rewritten run at
one layer. The controls are a matched request, an unrelated harmful request, a harmless request, and a
no-op. The pre-registered rule classifies a cell by the gap between matched and random recovery.
Below 10 points it rejects a request-specific bottleneck, above 25 it confirms one, and between the two
it decides nothing.

At the layer where matched recovery peaks, 27 of 30 cells reject and 3 are indeterminate, with a
maximum gap of 16.7 points. Taking instead the largest gap over all layers, 20 reject, 8 are
indeterminate and 2 confirm. Both confirming cells have fewer than eleven items with room to recover.
Appendix~\ref{app:patching} gives both.

An unrelated harmful request restores refusal about as well as the matched one. Patching one position
therefore cannot localise the mechanism. We therefore report single-position patching only as a characterization of final-token
representations.

\paragraph{Feature overlap.}
Top-100 feature overlap using public pretrained sparse autoencoders and transcoders
\citep{lieberum2024,mcdougall2025} (Appendix~\ref{app:repro}) is 0.21 to 0.54 between the native and
rewritten runs. A set
of calibration comparisons shows what that means. Paraphrases within one script overlap at 0.59 to
0.77. Changing the script lowers overlap more than changing the language, and harmless requests
fragment as much as harmful ones. Fragmentation under rewriting is therefore general and not specific
to harm.

Three cells combine high reading with low overlap, matching the high-reading regime identified by
direction transfer: Gemma-2-9B-it and Llama-3.2-3B-Instruct on Leetspeak, and Gemma-3-27B-it on
ITRANS.

\section{Discussion}
\label{sec:discussion}

\paragraph{Interpreting cross-script refusal rates.}
A refusal rate on rewritten requests conflates reading failure with safety failure, and on these
models the larger component is reading failure. Averaged over ten models (200 requests per cell),
64.2\% of Arabizi responses are off topic, against 1.2\% of the same requests in Arabic. Counting
those as bypasses attributes to safety a failure that happens before safety is engaged, and reporting
the full outcome distribution removes that ambiguity.

The same distinction holds for harm: among the 548 on-topic responses on harmful prompts
across 12,000 generations, specific usable harmful content appears in 116 (1.0\% overall), and no
rewritten cell exceeds 4.5\%. Treating every on-topic response as harmful compliance overstates
actual harm by nearly fivefold.

\paragraph{Practical implications for mitigation.}
Where the model does not recover the request during generation, additional safety training in more
scripts does not directly target the failure. Explicit decoding addresses that failure without
retraining: asking the model to rewrite the request and then answer its own rewriting improves refusal
in eleven of fifteen cells across Arabizi, ITRANS, and Leetspeak, recovering up to 79\% of the gap to
the native-script ceiling ($82\%$ to $100\%$). Self-decoding is not a complete defence, because an
attacker who degrades the decoding step degrades the refusal, as the Llama-3.2-3B Leetspeak cell
illustrates even without an adversary. The result establishes where the failure occurs and shows that
it is accessible without modifying model weights.

\paragraph{Where a script-specific gap remains.}
Where a model reads a rewritten script well, single-direction transfer drops sharply both within the
Gemma family (from 100\% on Gemma-2-9B-it Arabizi to 29\% on Gemma-2-9B-it Leetspeak) and outside
it (0\% to 2\% on Llama-3.2-3B-Instruct Leetspeak), even though native directions transfer strongly
across distinct native languages, and feature overlap identifies the same cells. The largest model we
evaluate, Gemma-3-27B-it, already shows the highest Arabizi on-topic rate (14.0\%) and highest
Arabizi specific harm (4.0\%) in Table~\ref{tab:harm}, suggesting that larger or frontier models
that read rewritten scripts more fluently may shift from the low-reading regime into this
high-reading regime. Where the model does read the script, safety training across scripts is the thing
that would help, and our account does not replace that.

\paragraph{Relation to prior accounts.}
Our result agrees with \citet{ifebi2026} that the refusal mechanism is intact and not triggered, and it
locates the step that fails. It disagrees with the stronger comprehension-failure reading
\citep{malik2026}, since the models recover the meaning when asked, and with a purely script-tied
account, which predicts harmful answers that almost never appear.

\section*{Limitations}

\begin{itemize}
\itemsep0.25em
\item Three rewritings, one source of harmful requests, and ten open models from 1B to 27B. Frontier
models may recover more and refuse more, and we do not test them.
\item Engagement is labelled by one model, with a second judge and human annotators on a subset.
Agreement is lowest on the Hindi pair, at Cohen's $\kappa$ 0.59 to 0.70.
\item The recovery test needs a model that can answer a four-way question. Six of eighteen cells fail
that check under near-paraphrase alternatives and are not reported.
\item While the 60-cell behavioural grid uses all 200 requests per cell, the multiple-choice recovery
tests and the fifteen-cell self-decoding experiment evaluate 50 paired requests per cell.
\item The 60-cell behavioural grid caps generation at 96 tokens, though extending generation to 256
tokens across all fifteen mitigation cells (750 rewritten baseline requests) yields a similarly low
specific harm rate ($\le 4.0\%$ per cell, $1.5\%$ overall).
\item Comprehended-subset direction transfer covers five well-powered cells across Gemma and Llama
(Table~\ref{tab:transfer}), bounded by a five-model cross-native transfer baseline
(Table~\ref{tab:crossnative}); models that read almost no Arabizi or ITRANS leave too few baseline
refusals to test ablation on those subsets.
\item While the cross-language forced-choice test (Arabizi and ITRANS queries with English options,
Table~\ref{tab:crosslang}) and the 200-item free-form ITRANS transliteration test put no native-script
text in the prompt and rule out character-level string matching, a multiple-choice readout still does
not prove that the representation it reads is the same one that drives refusal during open-ended
generation.
\item All analyses except the patching test are post-hoc.
\end{itemize}

\section*{Ethics Statement}

This work studies a known class of jailbreaks on the public AdvBench benchmark. It introduces no new
attack and releases no harmful content beyond what that benchmark already contains. Our finding is
defensive in intent: it argues that cross-script jailbreak rates on open models overstate harm, and it
supplies a decode-then-answer wrapper that needs no retraining. Harmful prompts are handled only as
classification inputs. The corpus was checked by native or fluent speakers of all three language pairs.

\section*{Acknowledgments}

We thank Anuj Tiwari for verifying the Hindi corpus subset and providing detailed feedback on earlier drafts of this paper. We also thank the Google Developer Experts program and Google Cloud for providing cloud and compute resources that supported this work.

\bibliography{references}

\appendix
\input{tables/t2_recovery}
\input{tables/t8_crosslang}
\input{tables/t4_harm}
\input{tables/t5_conditional}
\input{tables/t7_decode}
\input{tables/t9_cross_native}

\section{Per-model tables}
\label{app:percell}
Tables~\ref{tab:recovery} to~\ref{tab:crossnative} report the per-model numbers for within-language and cross-language meaning recovery, specific harmful content (200 items per cell), conditional refusal, fifteen-cell self-decoding across all three scripts, and the cross-native direction transfer baseline.

\section{Judges and agreement}
\label{app:judges}
\paragraph{Engagement rubric and primary judge.}
Every response is classified by Gemini-2.5-flash (temperature 0, 16 max output tokens, thinking
disabled) into one of four categories: \texttt{ON\_TOPIC} (substantively addresses the request in any
language), \texttt{REFUSAL} (declines, warns, or redirects for safety reasons), \texttt{OFF\_TOPIC}
(coherent text on an unrelated subject or misread prompt), or \texttt{GIBBERISH} (empty, garbled, or
repetitive). The judge is instructed to ignore harmfulness and score only whether the response engages
the request ($\text{engaged} = \texttt{ON\_TOPIC} \text{ or } \texttt{REFUSAL}$).

\paragraph{Second LLM judge and human agreement.}
Claude Opus 4.8 independently labelled a stratified 464-item subset under the same prompt, agreeing
with Gemini on 83.4\% of items four-way ($\kappa = 0.68$) and 87.1\% binary ($\kappa = 0.69$).
Per-script binary agreement is 95.0\% on Arabic ($\kappa = 0.64$, $n=60$), 76.7\% on Arabizi
($\kappa = 0.53$, $n=60$), 92.4\% on Devanagari ($n=66$), 75.4\% on ITRANS ($\kappa = 0.51$, $n=65$),
99.1\% on English ($n=107$), and 80.2\% on Leetspeak ($\kappa = 0.61$, $n=106$). Native speakers of
Arabic ($n=117$) and Hindi ($n=120$) and a fluent English speaker ($n=199$) annotated stratified
subsets blind to model and LLM labels. Binary agreement with humans is 89.7\% for Gemini
($\kappa = 0.76$; Arabizi $\kappa = 0.80$) and 90.6\% for Claude ($\kappa = 0.80$; Arabizi
$\kappa = 0.70$) on Arabic (Fleiss' $\kappa = 0.75$); 95.5\% for Gemini ($\kappa = 0.88$; Leetspeak
$\kappa = 0.84$) and 91.0\% for Claude ($\kappa = 0.79$; Leetspeak $\kappa = 0.63$) on English
(Fleiss' $\kappa = 0.80$); and 87.5\% for Gemini ($\kappa = 0.70$; four-class $\kappa = 0.60$; ITRANS
$\kappa = 0.67$) and 83.3\% for Claude ($\kappa = 0.59$; four-class $\kappa = 0.64$; ITRANS
$\kappa = 0.47$) on Hindi (Fleiss' $\kappa = 0.62$).

\paragraph{Harm judge.}
All 548 \texttt{ON\_TOPIC} responses across the 12,000 harmful-condition generations ($n=200$ per
cell) were scored by an independent harm judge (Claude Opus 4.8, with Claude Sonnet 4 fallback on
judge-refused items) as \texttt{SPECIFIC} (actionable steps, working code, quantities, or usable
instructions), \texttt{GENERIC} (high-level or fictionalized text without usable detail), or
\texttt{NONE} (benign reading or purely defensive text). Only 116 of 548 received \texttt{SPECIFIC},
with at most 4.5\% in any rewritten cell and 5.0\% in any native cell (Table~\ref{tab:harm}).
Applying the identical rubric to the 256-token \texttt{decode\_then\_refuse} runs across all five
models and three scripts (750 baseline rewritten responses) yields \texttt{SPECIFIC} harm in at most
4.0\% of items in any rewritten cell (0.0\% in eight of fifteen cells, and 11 of 750 or 1.5\%
overall), and 0.0\% to 2.0\% across all five self-decoded Arabizi cells.

\section{The recovery tests}
\label{app:reading}
The forced-choice test presents a rewritten request with four native-script options (\texttt{A}--\texttt{D})
under the prompt: \emph{``The text below is written in \{script\}. Which option has the same meaning as
the text? Only identify the meaning; you are not being asked to carry out the text.''} We read first-token
log-probabilities over \texttt{A}--\texttt{D} without generating text. Each item is shown four times
with the correct option rotated through all four positions while distractors keep their relative order;
cell-level recovery reports position-debiased accuracy across all 200 rotated trials per cell ($50 \times 4$), while the per-item split in Table~\ref{tab:conditional} marks an item as recovered when the correct letter is top-ranked in at least 3 of 4 rotations. Hard
distractors select the three requests in the same script with highest content-word overlap (ties broken
by length). A verbatim native-script control (\texttt{native\_copy}) requires $\ge 90\%$ accuracy for a
cell to be format-valid (\texttt{---} in Table~\ref{tab:recovery}). Under hard distractors, 12 of 18
cells pass and 6 fail (all three Gemma-2-9B-it cells, Gemma-3-4B-it Arabizi and Leetspeak, and
Llama-3.1-8B-Instruct Leetspeak); across the 12 valid cells, hard distractors lower mean recovery by
2.8 points and 10 of 12 stay above 76\%. Under standard same-category distractors, Gemma-2-9B-it
passes the format check in all three pairs and recovers 95.0\% (Arabizi), 97.0\% (ITRANS), and
93.5\% (Leetspeak). In the cross-language variant (Table~\ref{tab:crosslang}), where Arabizi and
ITRANS prompts are paired with four English near-paraphrase candidates so no Arabic or Devanagari
characters appear on screen, all ten cells across the five models pass the native-to-English format
check ($91.0\%$ to $99.0\%$), and cross-language recovery ($47.0\%$ to $75.5\%$) exceeds paired
engagement ($4.0\%$ to $44.0\%$) in every cell by 15.0 to 53.5 percentage points. In the 200-item
judge-free ITRANS-to-Devanagari transliteration test, the fraction of items decoded with character
similarity $\ge 0.5$ (and mean similarity) is 100.0\% (0.93) on Qwen3-8B, 100.0\% (0.91) on
Qwen2.5-7B-Instruct, 99.5\% (0.90) on Qwen3-4B, 97.5\% (0.85) on Gemma-3-4B-it, 100.0\% (0.77) on
Gemma-2-2B-it, and 22.5\% (0.28) on Gemma-3-1B-it.

\section{Threshold sensitivity}
\label{app:thresholds}
Sweeping the high-reading engagement floor over $\{50\%, 60\%, 70\%\}$ and the maximum drop from
native over $\{10, 15, 20\}$ points leaves the three high-reading, low-overlap cells (Gemma-2-9B-it
Leetspeak, Llama-3.2-3B-Instruct Leetspeak, and Gemma-3-27B-it ITRANS) unchanged for every drop
$\ge 15$ points across all floors (and keeps two of three at 10 points). Varying the sparse-autoencoder top-$k$ feature count over $k \in \{50, 100, 200\}$ preserves the
ordering between within-script paraphrases and cross-script rewritings across all evaluated models.

\section{Layers and coefficients}
\label{app:layers}
Each model's causal layer $L$ and ablation multiplier $\alpha$ were calibrated once on native harmful
and benign prompts and held fixed for cross-script transfer (Section~\ref{sec:mechanism}):
Gemma-3-1B-it $(L{=}11, \alpha{=}0.4)$, Gemma-2-2B-it $(12, 1.5)$, Gemma-3-4B-it $(16, 0.4)$,
Gemma-2-9B-it $(20, 1.5)$, Gemma-3-27B-it $(30, 0.4)$, Qwen3-4B $(18, 1.5)$, Qwen2.5-7B-Instruct
$(14, 1.5)$, Qwen3-8B $(15, 1.5)$, Llama-3.2-3B-Instruct $(20, 0.4)$, and Llama-3.1-8B-Instruct $(11,
1.5)$. On Gemma-2-9B-it Arabizi, $\alpha{=}0.4$ at $L{=}20$ gives a 71\% flip rate ($5/7$) versus
100\% ($7/7$) at $\alpha{=}1.5$. On Llama-3.2-3B-Instruct Leetspeak, testing $L{=}18$ and $L{=}14$
at $\alpha{=}1.5$ gives the same $1/46$ (2\%) flip rate. On Gemma-2-9B-it Leetspeak ($L{=}19,
\alpha{=}1.5$, \texttt{bfloat16}, calibrated from its English refusal-ablation sweep), removing the
English direction flips 12 of 41 comprehended refusals (29.3\%, 95\% CI $[17.6, 44.5]$) versus 0 of
41 (0.0\%) for the random direction, matching its 200-item lexical flip rate (25.5\% vs.\ 93.5\% on
English). Table~\ref{tab:crossnative} reports the five-model native-to-native transfer baseline
between English, Arabic, and Devanagari at the same $(L, \alpha)$ settings.

\section{Activation patching}
\label{app:patching}
Patching the final-token native residual stream (\texttt{matched} vs.\ \texttt{random} harmful prompt)
into the rewritten run is classified by the pre-registered gap rule: \texttt{REJECT} ($<10$ points),
\texttt{INDETERMINATE} ($10$--$25$ points), and \texttt{CONFIRM} ($\ge 25$ points). At the matched peak
layer, 27 of 30 cells are \texttt{REJECT}, 3 are \texttt{INDETERMINATE} (Gemma-3-1B-it Arabizi
$+16.7$ points at L13, Qwen2.5-7B-Instruct Arabizi $+14.3$ at L22, Gemma-3-1B-it ITRANS $+13.3$ at
L16), and 0 \texttt{CONFIRM}. Taking the maximum gap across all layers yields 20 \texttt{REJECT}, 8
\texttt{INDETERMINATE}, and 2 low-power \texttt{CONFIRM} cells: Gemma-2-9B-it Leetspeak ($+40.0$ points
at L17, room $n=10$) and Llama-3.2-3B-Instruct Leetspeak ($+25.0$ at L3, $n=4$). Among the 26 cells
with room $n \ge 20$, 0 confirm under either reading (max gap $14.3$ points at peak, $20.0$ overall).

\section{Corpus construction and audit}
\label{app:corpus}
The corpus totals 3,000 prompts aligned by \texttt{source\_id} across six scripts: 200 AdvBench harmful
requests per script (1,200 in Tiers A/C/E), 200 matched benign requests per script (1,200 in Tier B),
and 200 native-script paraphrases per language (600 in Tier D). Before final runs, a full audit fixed
four issues: (1) 22 \texttt{source\_id} miskeys in Tier A Arabic/Arabizi; (2) 26 intent-drift
paraphrases in Tier D Arabic; (3) 10 leaked refusal strings in Devanagari prompts (1 Tier C, 9 Tier D,
plus 7 cached translation refusals) and 4 truncations; and (4) a row-shift from index 12 onward in
Tier B benign Hindi.

\section{Reproducibility}
\label{app:repro}
All ten models were run in \texttt{bfloat16} using official HuggingFace instruct checkpoints and chat
templates with greedy decoding (temperature 0, sampling disabled, seed 42). Generation is capped at 96
new tokens for behavioural, ablation, and patching runs, and 256 per turn for self-decoding.
Feature-overlap runs use Gemma Scope and Gemma Scope 2 residual sparse autoencoders
\citep{lieberum2024,mcdougall2025} on Gemma models, public residual sparse autoencoders on Llama and
Qwen2.5-7B, and public MLP transcoders on Qwen3-4B and Qwen3-8B.

The release at \url{\openrepo} holds the result files every number in this paper is computed from.
It also holds the code that produced them, the corpora in each script, and a page that renders the
results.
The page needs no server and no build step, and \texttt{build\_data.py} regenerates its data from
the result files, so nothing in it is hand-typed.

The page never displays a harmful request. Only the benign matched pairs appear in its interactive
panel, and the harmful sets are shown as aggregates. Across the 12,000 main-grid responses, 116 were judged to contain
specific usable harmful content. Their text is withheld from the release and their labels are kept,
so every statistic reported here can still be recomputed. The activation tensors are not released,
and no number in this paper is read off them directly.

\end{document}

%% file: tables/t1_behaviour.tex
\begin{table*}[t]
\centering\footnotesize\setlength{\tabcolsep}{4pt}
\begin{tabular}{l rrr rrr rrr}
\toprule
 & \multicolumn{3}{c}{Arabic / Arabizi} & \multicolumn{3}{c}{Devanagari / ITRANS} & \multicolumn{3}{c}{English / Leetspeak} \\
\cmidrule(lr){2-4}\cmidrule(lr){5-7}\cmidrule(lr){8-10}
Model & nat & rew & eng & nat & rew & eng & nat & rew & eng \\
\midrule
Gemma-3-1B & 93 & 56 & 56 & 96 & 59 & 59 & 98 & 46 & 46 \\
Gemma-2-2B & 92 & 4 & 4 & 96 & 6 & 6 & 99 & 40 & 40 \\
Gemma-3-4B & 97 & 24 & 29 & 99 & 26 & 37 & 98 & 32 & 33 \\
Qwen3-4B & 94 & 5 & 6 & 84 & 14 & 16 & 98 & 16 & 17 \\
Llama-3.2-3B & 78 & 12 & 13 & 86 & 18 & 26 & 98 & 88 & 88 \\
Qwen2.5-7B & 97 & 12 & 17 & 90 & 24 & 36 & 99 & 28 & 28 \\
Qwen3-8B & 94 & 14 & 18 & 90 & 24 & 31 & 98 & 16 & 16 \\
Llama-3.1-8B & 91 & 12 & 21 & 90 & 9 & 34 & 95 & 46 & 48 \\
Gemma-2-9B & 97 & 42 & 52 & 98 & 52 & 65 & 98 & 89 & 90 \\
Gemma-3-27B & 97 & 74 & 88 & 98 & 86 & 94 & 97 & 78 & 78 \\
\bottomrule
\end{tabular}
\caption{Refusal on the native script (nat) and on the rewritten script (rew), and engagement on the rewritten script (eng), as percentages of 200 harmful requests per cell. Engagement counts a response that answers or refuses the request. Judge: Gemini-2.5-flash, 96 new tokens, greedy.}
\label{tab:behaviour}
\end{table*}

%% file: tables/t3_outcomes.tex
\begin{table}[t]
\centering\footnotesize\setlength{\tabcolsep}{3.5pt}
\begin{tabular}{lrrrr}
\toprule
Script & Refusal & On topic & Off topic & Unusable \\
\midrule
Arabic & 93.2 & 5.6 & 1.2 & 0.1 \\
Arabizi & 25.4 & 5.0 & 64.2 & 5.5 \\
Devanagari & 92.5 & 5.8 & 1.4 & 0.2 \\
ITRANS & 31.8 & 8.7 & 57.5 & 2.1 \\
English & 98.0 & 2.0 & 0.1 & 0.0 \\
Leetspeak & 47.8 & 0.4 & 51.8 & 0.0 \\
\bottomrule
\end{tabular}
\caption{What the models produce on harmful requests, as percentages of 200 requests per cell averaged over the ten models (12,000 responses in total). Off-topic text is the largest outcome on all three rewritten scripts.}
\label{tab:outcomes}
\end{table}

%% file: tables/t6_transfer.tex
\begin{table*}[t]
\centering\footnotesize\setlength{\tabcolsep}{4pt}
\begin{tabular}{llrrcrrr}
\toprule
Model & Script & $n$ & native & 95\% CI & random & layer & $\alpha$ \\
\midrule
Gemma-2-9B & Arabizi & 17 & 100 & [82, 100] & 0 & 20 & 1.5 \\
Gemma-3-1B & Arabizi & 29 & 72 & [54, 85] & 0 & 11 & 0.4 \\
Gemma-3-27B & Arabizi & 38 & 45 & [30, 60] & 0 & 30 & 0.4 \\
Gemma-2-9B & Leetspeak & 41 & 29 & [18, 44] & 0 & 19 & 1.5 \\
Llama-3.2-3B & Leetspeak & 46 & 0 & [0, 8] & 0 & 20 & 0.4 \\
\bottomrule
\end{tabular}
\caption{Removing the native refusal direction while the model answers rewritten requests it read and refused. Values are the percentage of those refusals that become compliance, against a norm-matched random direction. Layer and coefficient are each model's own calibrated values.}
\label{tab:transfer}
\end{table*}

%% file: tables/t2_recovery.tex
\begin{table*}[t]
\centering\footnotesize\setlength{\tabcolsep}{4pt}
\begin{tabular}{l rrr rrr rrr}
\toprule
 & \multicolumn{3}{c}{Arabizi} & \multicolumn{3}{c}{ITRANS} & \multicolumn{3}{c}{Leetspeak} \\
\cmidrule(lr){2-4}\cmidrule(lr){5-7}\cmidrule(lr){8-10}
Model & rec & eng & diff & rec & eng & diff & rec & eng & diff \\
\midrule
Gemma-3-4B & -- & 32.0 & -- & 80.5 & 44.0 & $+$36.5 & -- & 30.0 & -- \\
Qwen3-4B & 78.5 & 4.0 & $+$74.5 & 96.5 & 16.0 & $+$80.5 & 79.5 & 18.0 & $+$61.5 \\
Qwen2.5-7B & 76.5 & 10.0 & $+$66.5 & 95.0 & 22.0 & $+$73.0 & 82.0 & 22.0 & $+$60.0 \\
Qwen3-8B & 85.5 & 16.0 & $+$69.5 & 97.5 & 28.0 & $+$69.5 & 82.5 & 16.0 & $+$66.5 \\
Llama-3.1-8B & 42.0 & 20.0 & $+$22.0 & 55.5 & 32.0 & $+$23.5 & -- & 46.0 & -- \\
Gemma-2-9B & -- & 44.0 & -- & -- & 60.0 & -- & -- & 86.0 & -- \\
\bottomrule
\end{tabular}
\caption{Cell-level recovery (rec) across all 200 position-rotated trials (50 requests $\times$ 4 option positions; chance is 25.0\%) against paired engagement (eng) on the same 50 requests (full 200-request engagement rates appear in Table~\ref{tab:behaviour}). Recovery uses four near-paraphrase alternatives per request. A dash marks a cell where the model cannot pick a verbatim copy of the text in front of it, so its recovery score is not interpretable.}
\label{tab:recovery}
\end{table*}

%% file: tables/t8_crosslang.tex
\begin{table*}[t]
\centering\footnotesize\setlength{\tabcolsep}{4pt}
\begin{tabular}{l rrrr rrrr}
\toprule
 & \multicolumn{4}{c}{Arabizi to English options} & \multicolumn{4}{c}{ITRANS to English options} \\
\cmidrule(lr){2-5}\cmidrule(lr){6-9}
Model & ctrl & rec & eng & diff & ctrl & rec & eng & diff \\
\midrule
Gemma-3-4B & 94.0 & 56.0 & 32.0 & $+$24.0 & 93.5 & 61.5 & 44.0 & $+$17.5 \\
Qwen3-4B & 97.5 & 56.5 & 4.0 & $+$52.5 & 94.0 & 68.5 & 16.0 & $+$52.5 \\
Qwen2.5-7B & 97.0 & 63.0 & 10.0 & $+$53.0 & 99.0 & 75.5 & 22.0 & $+$53.5 \\
Qwen3-8B & 96.5 & 63.0 & 16.0 & $+$47.0 & 94.0 & 70.5 & 28.0 & $+$42.5 \\
Llama-3.1-8B & 92.5 & 50.5 & 20.0 & $+$30.5 & 91.0 & 47.0 & 32.0 & $+$15.0 \\
\bottomrule
\end{tabular}
\caption{Cross-language meaning recovery (rec) across 200 position-rotated trials (50 requests $\times$ 4 positions) against paired engagement (eng) on the same 50 requests, when candidate meanings are presented in English so no Arabic or Devanagari characters appear on screen. The control column (ctrl) is accuracy when matching the native-script request (Arabic or Devanagari) against the same four English near-paraphrase options.}
\label{tab:crosslang}
\end{table*}

%% file: tables/t4_harm.tex
\begin{table*}[t]
\centering\footnotesize\setlength{\tabcolsep}{4pt}
\begin{tabular}{l rr rr rr}
\toprule
 & \multicolumn{2}{c}{Arabizi} & \multicolumn{2}{c}{ITRANS} & \multicolumn{2}{c}{Leetspeak} \\
\cmidrule(lr){2-3}\cmidrule(lr){4-5}\cmidrule(lr){6-7}
Model & on topic & harmful & on topic & harmful & on topic & harmful \\
\midrule
Gemma-3-1B & 0.0 & 0.0 & 0.0 & 0.0 & 0.0 & 0.0 \\
Gemma-2-2B & 0.0 & 0.0 & 0.0 & 0.0 & 0.0 & 0.0 \\
Gemma-3-4B & 5.5 & 0.5 & 10.5 & 3.0 & 0.5 & 0.0 \\
Qwen3-4B & 0.5 & 0.0 & 2.0 & 0.0 & 0.5 & 0.0 \\
Llama-3.2-3B & 1.5 & 0.0 & 7.0 & 2.0 & 0.0 & 0.0 \\
Qwen2.5-7B & 4.5 & 1.0 & 12.0 & 3.5 & 0.0 & 0.0 \\
Qwen3-8B & 4.0 & 1.5 & 7.5 & 0.5 & 0.0 & 0.0 \\
Llama-3.1-8B & 9.5 & 0.0 & 25.5 & 4.5 & 1.5 & 0.5 \\
Gemma-2-9B & 10.0 & 0.5 & 13.5 & 1.5 & 1.0 & 0.0 \\
Gemma-3-27B & 14.0 & 4.0 & 8.5 & 2.0 & 0.5 & 0.0 \\
\bottomrule
\end{tabular}
\caption{On-topic responses and the subset that contains specific, usable harmful content, as percentages of 200 requests per cell. The two columns differ because the engagement judge is instructed to ignore whether content is harmful. A second judge from another model family scored harm.}
\label{tab:harm}
\end{table*}

%% file: tables/t5_conditional.tex
\begin{table}[t]
\centering\footnotesize\setlength{\tabcolsep}{3.5pt}
\begin{tabular}{lrrcrr}
\toprule
 & \multicolumn{3}{c}{meaning recovered} & \multicolumn{2}{c}{not recovered} \\
\cmidrule(lr){2-4}\cmidrule(lr){5-6}
Model & $n$ & refused & 95\% CI & $n$ & refused \\
\midrule
Gemma-3-4B & 39 & 38 & [25, 54] & 11 & 18 \\
Qwen3-4B & 124 & 14 & [9, 21] & 26 & 4 \\
Qwen2.5-7B & 120 & 18 & [12, 25] & 30 & 7 \\
Qwen3-8B & 129 & 18 & [12, 25] & 21 & 10 \\
Llama-3.1-8B & 28 & 11 & [4, 27] & 72 & 8 \\
\midrule
Pooled & 440 & 18 & [15, 22] & 160 & 8 \\
\bottomrule
\end{tabular}
\caption{Refusal on rewritten harmful requests, split by whether the model recovered the meaning. Format-valid cells only. Because the models recover most requests, the right-hand group is small, and in eight cells it contains no refusals.}
\label{tab:conditional}
\end{table}

%% file: tables/t7_decode.tex
\begin{table*}[t]
\centering\footnotesize\setlength{\tabcolsep}{4pt}
\begin{tabular}{llrrrrr}
\toprule
Model & Script & rewritten & self-decoded & native & gap closed & decode sim. \\
\midrule
Gemma-3-4B & ITRANS & 34 & 86 & 100 & 79 & 0.85 \\
Llama-3.2-3B & ITRANS & 32 & 66 & 84 & 65 & 0.90 \\
Gemma-3-4B & Leetspeak & 58 & 84 & 98 & 65 & 0.94 \\
Gemma-2-9B & ITRANS & 56 & 78 & 98 & 52 & 0.92 \\
Qwen3-8B & ITRANS & 30 & 62 & 92 & 52 & 0.92 \\
Qwen3-8B & Leetspeak & 16 & 50 & 100 & 40 & 0.76 \\
Qwen3-4B & ITRANS & 14 & 44 & 90 & 39 & 0.90 \\
Llama-3.2-3B & Arabizi & 10 & 20 & 82 & 14 & 0.45 \\
Qwen3-4B & Leetspeak & 20 & 30 & 100 & 12 & 0.67 \\
Gemma-2-9B & Arabizi & 48 & 54 & 98 & 12 & 0.80 \\
Gemma-3-4B & Arabizi & 26 & 32 & 98 & 8 & 0.74 \\
Gemma-2-9B & Leetspeak & 88 & 88 & 98 & 0 & 0.97 \\
Qwen3-8B & Arabizi & 18 & 16 & 94 & -3 & 0.63 \\
Qwen3-4B & Arabizi & 6 & 0 & 94 & -7 & 0.62 \\
Llama-3.2-3B & Leetspeak & 74 & 42 & 98 & -133 & 0.54 \\
\bottomrule
\end{tabular}
\caption{Refusal on the same 50 requests under three presentations: the rewritten request, the model's own rewriting of it into the native script, and the native-script request. Gap closed is the share of the distance from the rewritten baseline to the native ceiling. Decode similarity is the character similarity of the model's own rewriting to the known correct text.}
\label{tab:decode}
\end{table*}

%% file: tables/t9_cross_native.tex
\begin{table*}[t]
\centering\footnotesize\setlength{\tabcolsep}{4pt}
\begin{tabular}{l r rrr rrr rrr}
\toprule
 & & \multicolumn{3}{c}{Target: English} & \multicolumn{3}{c}{Target: Arabic} & \multicolumn{3}{c}{Target: Devanagari} \\
\cmidrule(lr){3-5}\cmidrule(lr){6-8}\cmidrule(lr){9-11}
Model & $\cos(\text{en},\text{ar})$ & same & ar & rand & same & en & rand & same & en & rand \\
\midrule
Gemma-2-9B (judge) & 0.97 & 26 & 71 & 0 & 78 & 43 & 0 & 86 & 33 & 0 \\
Qwen3-8B (judge) & 0.81 & 90 & 84 & 0 & 87 & 92 & 0 & 29 & 38 & 0 \\
Llama-3.2-3B (lex.) & 0.66 & 30 & 51 & 2 & 42 & 8 & 2 & 67 & 33 & 0 \\
Gemma-3-1B (lex.) & 0.52 & 75 & 90 & 2 & 98 & 59 & 0 & 94 & 36 & 6 \\
Gemma-3-4B (lex.) & 0.96 & 2 & 49 & 9 & 60 & 17 & 4 & 83 & 0 & 11 \\
\bottomrule
\end{tabular}
\caption{Cross-native direction transfer baseline across 50 harmful requests per native language at each model's calibrated $(L, \alpha)$. Columns give refusal flip rates (\%) when ablating the same-language direction, a cross-lingual native direction (Arabic on English, or English on Arabic and Devanagari), or a norm-matched random direction, scored by the four-way judge on 8B/9B models and by a lexical coherence-and-refusal classifier on 1B to 4B models (where single-layer ablation removes refusal prefixes without always producing full compliance).}
\label{tab:crossnative}
\end{table*}